\documentclass[10pt]{article} 
\usepackage[usenames,dvipsnames,table]{xcolor}

\usepackage{tmlr}

\usepackage{amsmath,amsfonts,bm}

\def\eqref#1{equation~\ref{#1}}

\def\1{\bm{1}}

\DeclareMathAlphabet{\mathsfit}{\encodingdefault}{\sfdefault}{m}{sl}
\SetMathAlphabet{\mathsfit}{bold}{\encodingdefault}{\sfdefault}{bx}{n}

\usepackage{url}
\usepackage{amsmath}
\usepackage{amssymb}
\usepackage{mathtools}
\usepackage{graphicx}
\usepackage[colorlinks=true, allcolors=blue]{hyperref}
\usepackage{algpseudocode}
\usepackage{algorithm}
\usepackage{setspace}
\usepackage{booktabs}
\usepackage{subcaption}
\usepackage{tabularray}
\usepackage{hyperref}
\usepackage{tikz}
\usepackage{multirow}
\usepackage{booktabs}
\usepackage[capitalise]{cleveref}
\usetikzlibrary{bayesnet, calc, arrows.meta, bending}
\usetikzlibrary{decorations.pathreplacing,angles,quotes,shapes.multipart, calligraphy}

\newcommand{\interv}[1]{\textcolor{RoyalBlue}{#1}}
\newcommand{\unint}[1]{\textcolor{Bittersweet}{#1}}
\definecolor{graybg}{gray}{1.0}

\title{Mitigating attribute amplification in counterfactual image generation}

\author{\name Tian Xia \email t.xia@imperial.ac.uk \\
      \addr Imperial College London, UK
      \AND
      \name Mélanie Roschewitz  \email m.roschewitz21@imperial.ac.uk \\
      \addr Imperial College London, UK
      \AND
      \name Fabio De Sousa Ribeiro \email f.de-sousa-ribeiro@imperial.ac.uk \\
      \addr Imperial College London, UK
      \AND
      \name Charles Jones
      \email charles.jones17@imperial.ac.uk \\
      \addr Imperial College London, UK
      \AND
      \name Ben Glocker
      \email b.glocker@imperial.ac.uk \\
      \addr Imperial College London, UK
      }

\begin{document}

\maketitle

\begin{abstract}
Causal generative modelling is gaining interest in medical imaging due to its ability to answer interventional and counterfactual queries. Most work focuses on generating counterfactual images that look plausible, using auxiliary classifiers to enforce \textit{effectiveness} of simulated interventions. We investigate pitfalls in this approach, discovering the issue of \textit{attribute amplification}, where unrelated attributes are spuriously affected during interventions, leading to biases across protected characteristics and disease status. We show that attribute amplification is caused by the use of hard labels in the counterfactual training process and propose \textit{soft counterfactual fine-tuning} to mitigate this issue. Our method substantially reduces the amplification effect while maintaining effectiveness of generated images, demonstrated on a large chest X-ray dataset. Our work makes an important advancement towards more faithful and unbiased causal modelling in medical imaging.
\end{abstract}

\section{Introduction}



Scientific investigation has always been driven by causal questions such as: \textit{``What is the effect of treatment X on disease Y?''}. In medical imaging, we may ask \textit{``What would this patient's image look like if they had no disease?''}. These causal questions cannot be answered with statistical tools alone, but require a mathematical framework that allows such questions to be answered from data. Causal models describe (assumed) causal mechanisms of a system~\citep{pearl2009causality}, in which the causal relationships between variables are directed from cause to effect, and change in a cause would result in change in its effect, but not the other way around. Causal models allow us to analyse interactions of variables within our environment (interventions) and hypothetical alternative worlds (counterfactuals). The ability to reason about cause-and-effect relationships has gained significant interest~\citep{scholkopf2021toward,scholkopf2022causality,zhang2023towards}. Efforts have been made to combine causality and deep learning models~\citep{Bengio2013,scholkopf2021toward}, but few works have attempted to satisfy all three rungs of Pearl's causal ladder~\citep{pearl2009causality}: (i) association, (ii) intervention and (iii) counterfactual. Notable works include Deep Structural Causal Models (DSCMs)~\citep{pawlowski2020deep,de2023high} and Neural Causal Models (NCMs)~\citep{xia2021causal,xia2023neural}. Our work builds upon DSCMs used for producing high-fidelity counterfactual images for real-world data~\citep{de2023high}. \citet{de2023high} proposed to train the generative causal model using a hierarchical variational auto-encoder (HVAE) conditioned on the assumed causal parents. Solely relying on standard likelihood-based training, however, was found to lead to suboptimal axiomatic \textit{effectiveness}~\citep{monteiro2023measuring}, where conditioning on intervened upon parents may be ignored by the forward model post-abduction. To encourage the model to obey (counterfactual) conditioning, the authors proposed an additional training step called \textit{counterfactual fine-tuning}, whereby pretrained parent predictors are used to help fine-tune the HVAE such that conditioning on counterfactual parents results in semantically meaningful changes.

We analyse the quality of counterfactuals generated by DSCMs under the counterfactual fine-tuning setup. We discover that while DSCMs produce plausible counterfactual images, these may exhibit amplified attributes that were not intervened upon. For example, when intervening on biological sex for a healthy male patient, a generated female counterfactual may appear healthier than the real image. We term this phenomenon \textit{attribute amplification}.
Amplification of unrelated attributes needs to be addressed as it is in conflict with the assumed causal graph and can cause distribution shift and introduce harmful bias. Attribute amplification can lead to the encoding of spurious correlations in the generated images between protected characteristics and disease.
We find that attribute amplification occurs in the counterfactual fine-tuning step. To mitigate this issue, we propose a simple yet effective strategy, \textit{soft counterfactual fine-tuning}, where we replace hard labels with inferred \textit{soft} labels during counterfactual fine-tuning. We show through a series of experiments that soft counterfactual fine-tuning effectively mitigates attribute amplification in counterfactual inference models, whilst retaining the ability of high-fidelity image synthesis.

\section{Counterfactual image generation}

Various works have used generative models such as VAEs~\citep{kingma2013auto}, GANs~\citep{goodfellow2020generative}, normalizing flows~\citep{rezende2015variational} and diffusion models~\citep{sohl2015deep,ho2020denoising,song2021scorebased} for causal effect estimation~\citep{louizos2017causal,kocaoglu2017causalgan,tran2017implicit}, causal discovery~\citep{yang2021causalvae,sanchez2022diffusion,geffner2022deep}, and other tasks modelling conditional~\citep{trippe2018conditional,mirza2014conditional,sohn2015learning,dhariwal2021diffusion} and interventional distributions~\citep{kocaoglu2018causalgan,ke2019learning,xia2021causal,zevcevic2021relating}. Very few works~\citep{xia2023neural,pawlowski2020deep} satisfied all three rungs of Pear's ladder of causation~\citep{pearl2009causality,bareinboim2022pearl}. Our work builds upon DSCMs first introduced by \citet{pawlowski2020deep}, 
and recently improved by \citet{de2023high}, leveraging hierarchical variational autoencoders (HVAE) to generate high-quality high-resolution images. In the following, we summarise the main components of DSCMs as introduced by Ribeiro et al., {for more details refer to~\citet{de2023high}}.



\paragraph{Structural Causal Models}\citep{pearl2009causality} are defined by a triplet $\langle U, A, F \rangle$, where $U=\{u_{i}\}_{i=1}^K$ are a set of \textit{exogenous variables}, $A=\{a_{i}\}_{i=1}^K$ a set of \textit{endogenous variables}, and $F=\{f_{i}\}_{i=1}^K$ a set of functions such that $a_{k}:=f_{k}(\mathbf{pa}_{k},u_{k})$,
where $\mathbf{pa}_{k}\subseteq A\setminus a_{k}$ are called direct causes or \textit{parents} of $a_{k}$. SCMs allow us to perform interventions by substituting one or multiple parents, denoted by the \textit{do}-operator. The estimation of counterfactuals follows three steps: (i) \textit{abduction}: infer exogenous noise given observed data; (ii) \textit{action}: perform an intervention  $do(a_{k}:=c)$; (iii) \textit{prediction}: infer counterfactuals via the modified model.

\paragraph{Deep Structural Causal Models} were first introduced in \citet{pawlowski2020deep} and recently improved in \citet{de2023high} for high-resolution image generation. Let $\mathbf{x}$ be an image and $\{a_{1}, ..., a_{K-1} \} \supseteq \mathbf{pa_{x}}$ \textit{ancestors} of $\mathbf{x}$. For each low-dimensional attribute, a conditional normalizing flow is used as its (invertible) mechanism $a_{k}=f_{k}(u_{k}; \mathbf{pa}_{k})$, such that abduction is tractable and explicit. For high-dimensional structured variables like images, the mechanism is implemented by a HVAE. To generate a counterfactual, we first approximately infer the exogenous noise $\mathbf{z}\sim q_{\phi}(\mathbf{z} \mid \mathbf{x}, \mathbf{pa}_{\mathbf{x}})$, where $q_{\phi}$ is the encoder of the HVAE. Similarly, we can explicitly infer the exogenous noises for attributes via: 
$u_{k}=f^{-1}(a_{k}; \mathbf{pa}_{k})$. We can then perform an intervention $do(a_i\coloneqq c)$ by e.g. setting $a_{i}$ to some target $c$. Note, we may intervene on multiple $a_{i}$ at the same time. We compute counterfactuals of attributes using abducted noise $u_{k}$, and thus obtain $\mathbf{\widetilde{pa}_{x}}$, the counterfactuals of parents of $\mathbf{x}$. Finally, we generate the counterfactual image $\widetilde{\mathbf{x}}=g_{\theta}(\mathbf{z}, \mathbf{\widetilde{pa}_{x}})$. \citet{de2023high} noticed that using only likelihood training for the HVAE, some parents may be ignored during inference, effectively leading to `ignored counterfactual conditioning', i.e. $\mathbf{\widetilde{x}}$ does not obey $\mathbf{\widetilde{pa}_{x}}$. To mitigate this issue the authors introduced \textit{counterfactual fine-tuning} using hard labels for parents. \emph{Hard-CFT} leverages a pre-trained predictor $q_{\psi}(\mathbf{\widetilde{pa}_{x}} \mid \widetilde{\mathbf{x}})$, and optimises the pre-trained HVAE weights $\{\theta,\phi\}$ to maximise $\log q_{\psi}(\mathbf{\widetilde{pa}_{x}} \mid \widetilde{\mathbf{x}})$ with $\psi$ fixed. This encourages the DSCM to produce effective counterfactuals which obey counterfactual conditioning.

\section{The attribute amplification problem}
\label{sec: attribute amplfication}
%
\paragraph{Datasets and experimental setup.}
We use chest X-ray images of the MIMIC-CXR dataset~\citep{johnson2019mimic} for our experiments. We follow the dataset splits and filtering of \citep{de2023high,glocker2023algorithmic}, and focus on the disease label of pleural effusion. The resulting dataset contained subjects with and without pleural effusion, consisting of 62,336 images for training, 9,968 for validation and 30,535 for testing. All images were resized to 224×224 resolution. Following \citet{de2023high}, we consider four attributes: (self-reported) race, biological sex, disease status, and age for the underlying causal graph (see \cref{fig:amplification violate causal graph}). Note, that disease, sex, and race are assumed to be independent.

\paragraph{Evaluating counterfactual effectiveness.} The ability to generate realistic counterfactual images can be measured via \textit{effectiveness}~\citep{de2023high,monteiro2023measuring}, which assesses whether counterfactuals (CFs) can fool an attribute predictor trained on real images. Here, we measure the predictive performance in terms of AUC. Results are summarised in \cref{tab:effectiveness}, where we report (i) the baseline test performance of each attribute predictor on real images and (ii) the performance when testing on different types of CFs. Ideally, we would find similar performance across real and counterfactual images. Without counterfactual fine-tuning, CFs yield much lower AUC on intervened attributes, highlighting the importance of CFT.

\paragraph{The attribute amplification problem.} From the results in \cref{tab:effectiveness}, we identify the key problem of \emph{attribute amplification}, where attributes that should remain unchanged during interventions are amplified in the generated images, indicated by an increase in AUC beyond the baseline for the attribute predictor. When using Hard-CFT as proposed in \citet{de2023high}, counterfactual images yield consistently higher AUC compared to real images for \textit{unintervened} attributes. These attributes seem more strongly encoded in the generated images than in the real images, leading to undesired dataset shift and potential bias. For example, if $x$ is \textit{healthy} and \textit{male}, when $do(sex:=female)$, $\widetilde{\mathbf{x}}$ becomes \textit{healthier} than $\mathbf{x}$, although in the assumed causal graph $disease$ is not a descendant of $sex$ and thus $do(sex)$ should not affect $disease$. We observe an increase in AUC for the disease attribute by 3.6$\%$ after intervening on sex. Similar, the AUC for race increases by 3.5$\%$ after intervening on disease. These undesirable side effects question the faithfulness of generated counterfactual images when using the Hard-CFT strategy.

\paragraph{Why is attribute amplification a problem?} The effect on unintervened attributes violates the assumed causal graph, affecting the causal relationships between variables. This violation is illustrated in \cref{fig:amplification violate causal graph}. Attribute amplification not only affects the quality of counterfactuals but may introduce harmful, spurious correlations between protected characteristics and disease status. This could have negative consequences when using generated images in downstream applications such as counterfactual data augmentation or counterfactual explainability.

\begin{table}[b!]
\centering
\caption{Quantitative evaluation of \textit{effectiveness} of CFs of different types. For \interv{intervened} and \unint{unintervened} attributes, AUC $\%$ (test on CFs) is reported (with change in brackets) compared to the baseline (test on real images, first row).
}
\label{tab:effectiveness}
\begin{tabular}{@{}lp{0.03\textwidth}lp{0.02\textwidth}cp{0.03\textwidth}cp{0.03\textwidth}c@{}}
\toprule
\multirow{2}{*}{\textsc{Intervention}} && \multirow{2}{*}{\textsc{CF Type}} && \multicolumn{5}{c}{\textsc{\interv{Effectiveness} / \unint{Amplification}}} \\
\cmidrule{5-9}
  && && \textsc{Race}        && \textsc{Sex}         && \textsc{Disease}     \\ 
\midrule
None   &&       Real images    && 93.8        && 99.6        && 94.2            \\

\midrule
\multirow{3}{*}{$do(\text{race})$}  && No CFT    && \cellcolor{graybg}  \interv{68.9  \small{($\downarrow$  24.9)}}                   && \unint{{99.6 \small{($\cdot$ 0.0)}}}                        && \unint{{94.2  \small{($\cdot$ 0.0)}}}                                       \\
                        && Hard-CFT       && \cellcolor{graybg}  \interv{{99.0 \small{($\uparrow$  5.2)}}}                       && \unint{99.7 \small{($\uparrow$  0.1)}}                        && \unint{96.4 \small{($\uparrow$  2.2)}}                                         \\
                        && Soft-CFT      && \cellcolor{graybg} \interv{98.7 \small{($\uparrow$  4.9)}}                       && \unint{99.5 \small{($\downarrow$  0.1)}}                       && \unint{94.3 \small{($\uparrow$  0.1)}}                                          \\
\midrule
\multirow{3}{*}{$do(\text{sex})$}  && No CFT     && \unint{92.4  \small{($\downarrow$  1.4)}}                        && \cellcolor{graybg} \interv{92.8 \small{($\downarrow$ 6.8)}}                       && \unint{{94.0 \small{($\downarrow$ 0.2)}}}                                        \\
                        && Hard-CFT       && \unint{97.1 \small{($\uparrow$ 3.3)}}                       && \cellcolor{graybg} \interv{{99.8 \small{($\uparrow$ 0.2)}}}                        && \unint{97.8 \small{($\uparrow$  3.6)}}                                         \\
                        && Soft-CFT      && \unint{{93.3}  \small{($\downarrow$  0.5)}}                         && \cellcolor{graybg} \interv{99.7  \small{($\uparrow$  0.1)}}                      && \unint{94.5  \small{($\uparrow$  0.3)}}                                       \\
\midrule
\multirow{3}{*}{$do(\text{disease})$} && No CFT     &&  \unint{93.7  \small{($\downarrow$ 0.1)}}                        && \unint{99.2  \small{($\downarrow$ 0.4)}}                       && \cellcolor{graybg} \interv{70.6  \small{($\downarrow$ 23.6})}                                      \\
                        && Hard-CFT       &&  \unint{97.3  \small{($\uparrow$ 3.5)}}                        && \unint{99.7  \small{($\uparrow$ 0.1)}}                        && \cellcolor{graybg} \interv{97.9  \small{($\uparrow$ 3.7)}}                                         \\
                        && Soft-CFT      && \unint{{93.8}  \small{($\cdot$ 0.0)}}                       && \unint{{99.6}  \small{($\cdot$ 0.0)}}                        && \cellcolor{graybg} \interv{{98.1}  \small{($\uparrow$ 3.9)}}                        \\     
\bottomrule
\end{tabular}
\end{table}

\paragraph{Why does Hard-CFT cause attribute amplification?}  
In Hard-CFT, $\mathbf{\widetilde{x}}$ is encouraged to maximise the performance of a pretrained attribute predictor $q_{\psi}(\widetilde{a}_{k}\mid\mathbf{\widetilde{x}}), \forall\widetilde{a}_{k} \in \mathbf{\widetilde{pa}_{x}}$, regardless of whether $\widetilde{a}_{k}$ should actually be affected by the intervention. Suppose we have sample $\mathbf{x}$ with mild pleural effeusion ($d=1$), and that the attribute predictor outputs a disease probability of $q_\psi(d\mid\mathbf{x})=0.65$ for the real image. When we intervene on race, as both race and disease are parents of $\mathbf{x}$, Hard-CFT will not only maximise the probability of the race predictor but will also maximise the predicted disease probability $q_{\psi}(\widetilde{d} \mid \mathbf{\widetilde{x}})$ by optimizing the HVAE weights. As a consequence, the model will amplify pleural effusion related features in the generated race counterfactual image.

\section{Mitigating attribute amplification with soft labels}

\begin{algorithm}[!t]
\caption{Soft counterfactual fine-tuning (Soft-CFT)}
\label{alg: Soft-CFT}
\begin{algorithmic}
\renewcommand{\baselinestretch}{1.09}\selectfont
\State {\textbf{Input:}  training data $\{ \mathbf{x}, \mathbf{pa_{x}} \}$; frozen predictor $q_{\psi}$; DSCM $q_{\phi}$, $g_{\theta}$.}
\State {\textbf{Compute counterfactuals:}} 
\State \indent {1. Compute $\mathbf{\widetilde{x}}$ and $\mathbf{\widetilde{pa}_{x}}$ using $q_{\phi}$ and $g_{\theta}$ upon intervention $\mathbf{do(\cdot)}$.}
\State {\textbf{Fine-tune DSCM:}}
\State \indent {1. Identify attributes in $\mathbf{\widetilde{pa}_{x}}$ that are not intervened on or non-descendants of intervened attributes, \indent \hspace{7pt}  denoted as $\mathbf{A^{*}}$.} 
\State \indent {2. For attributes $a_{k} \in \mathbf{A^{*}}$ use soft labels as targets.  With $l$ a classification  loss function: $\mathcal{L}_{soft}:=$ \indent \hspace{7.5pt} $l(q_{\psi}(a_{k} \mid \mathbf{{x}}), q_{\psi}(a_{k} \mid \mathbf{\widetilde{x}}))$.}
\State \indent {4. For ${a}_{k} \notin \mathbf{A^{*}}$ use hard labels as targets i.e. $\mathcal{L}_{interv}:=l({a}_{k}, q_{\psi}({a}_{k} \mid \mathbf{{\widetilde{x}}}))$. 
\State \indent {5. Fine-tune $q_{\phi}$, $g_{\theta}$ by minimising $\mathcal{L}_{soft}+\mathcal{L}_{interv}$.} 
}
\end{algorithmic}
\end{algorithm}

We discovered that attribute amplification occurs during counterfactual fine-tuning, which, however, is necessary to increase effectiveness on intervened attributes. Amplification of unintervened attributes is caused by the use of hard labels in the CFT step, making all attributes more extreme than they may have been in the original image. Ideally, we only change attributes that are affected by the intervention, and leave other attributes unchanged. To this end, we propose soft counterfactual fine-tuning (Soft-CFT) to mitigate attribute amplification. In Soft-CFT, we treat attributes differently during fine-tuning depending on whether they are being intervened on or not. For intervened attributes, we fine-tune the model with the hard labels to ensure the model obeys conditioning. For unintervened attributes, we use inferred soft labels. We first pass the real image through the trained attribute predictors to obtain the predicted probability for each non-intervened attribute. These predicted probabilities then form the targets during counterfactual fine-tuning. This encourages the network to not change the encoding of unintervened attributes, and the resulting counterfactuals should have the same predicted probability on the unintervened attributes as the original real images. The Soft-CFT process is summarised in \cref{alg: Soft-CFT}.

\subsection{Effect of Soft-CFT on effectiveness and attribute amplification}

\begin{figure*}[t!]
    \centering
    \begin{subfigure}{0.566\textwidth}
    \centering
    \includegraphics[width=\textwidth]{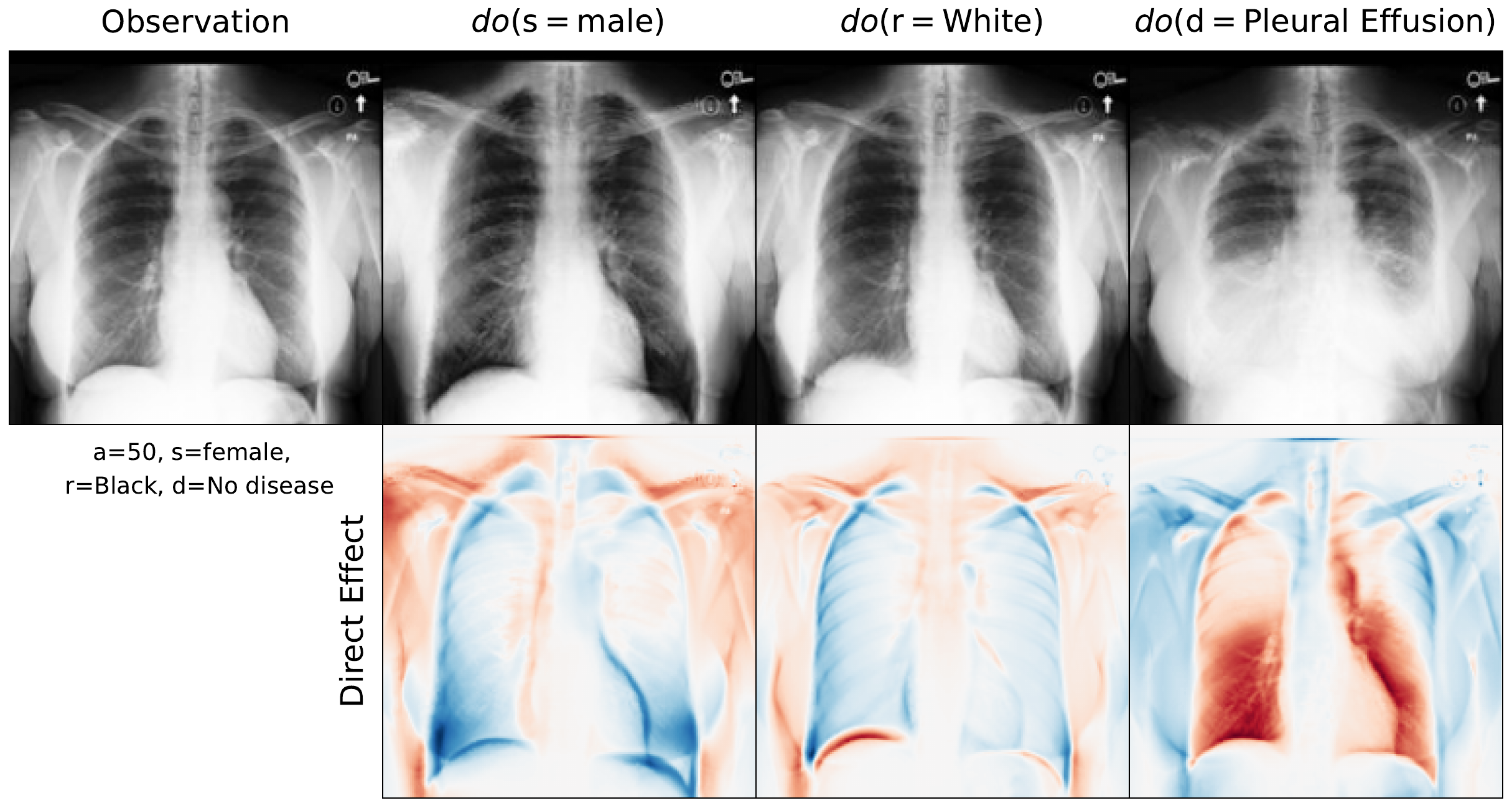}
    \caption{Hard-CFT}
    \label{fig:cft cf}
    \end{subfigure}
    \begin{subfigure}{0.424\textwidth}
    \centering
    \includegraphics[width=\textwidth,trim={10.35cm 0 0 0},clip]{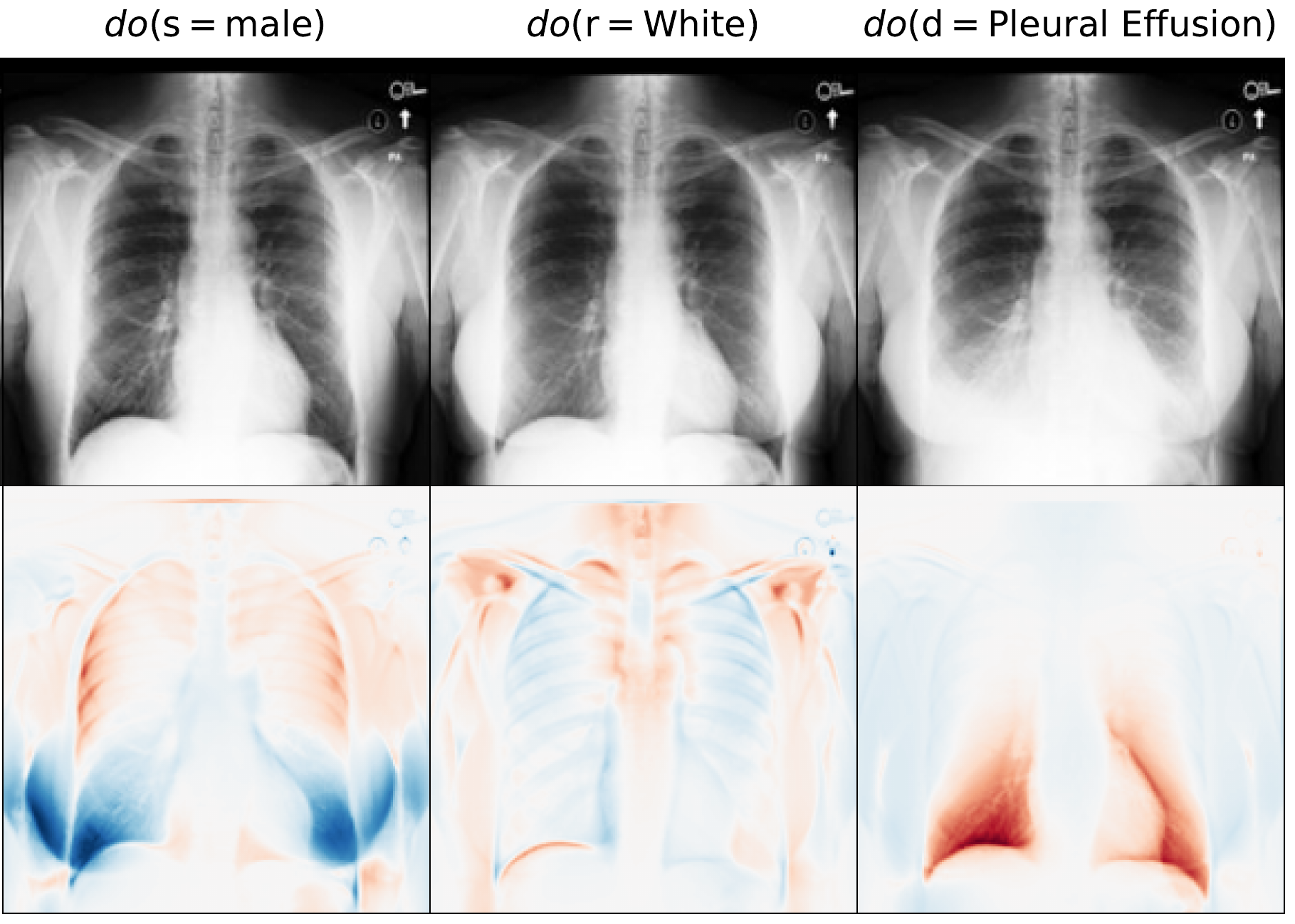}
    \caption{Soft-CFT}
    \label{fig:Soft-CFT cf}
    \end{subfigure}
    \caption{Generated CFs with (a) Hard-CFT and (b) Soft-CFT. First rows show original image $\mathbf{x}$ and CFs $\mathbf{\widetilde{x}}$; second rows show direct effect of CFs, i.e. $\mathbf{\widetilde{x}}-\mathbf{x}$.}
    \label{fig:CFs}
\end{figure*}

Table \ref{tab:effectiveness} compares effectiveness and amplification of Soft-CFT and Hard-CFT. With Soft-CFT, AUCs for unintervened attributes are closer to the baseline, substantially reducing the amplification effect while preserving effectiveness on intervened attributes. When intervening on sex, disease amplification is reduced to 0.3$\%$ compared to 3.6$\%$ for Hard-CFT. Amplification on race is completely removed (0.0) compared to 3.5$\%$ when intervening on disease. Visual examples in \cref{fig:CFs} show that both Hard-CFT and Soft-CFT generate plausible CFs. However, with Hard-CFT, the generated images exhibit more global and stronger direct effects. With Soft-CFT, images show more targeted, localised changes aligned with the intervened variable. See \cref{fig:more CFs} for additional visual examples.

\subsection{Utility of counterfactual images in downstream applications}

We evaluate the utility of CFs by training multiple attribute predictors on the generated counterfactuals. We first generate all possible counterfactuals for every attribute and every image in the training set. We then train attribute predictors on each set of counterfactuals (race, sex, disease), and evaluate on real images from the test set. Note, this is the opposite in \cref{tab:effectiveness} where attribute predictors are trained on real images, and evaluated on counterfactuals. To compare the performance of predictors trained on counterfactuals only, we compare to the baseline where predictors are trained and tested on real data only. Without CFT, AUCs are generally low on intervened attributes due to the low effectiveness (see \cref{tab:effectiveness}). We find consistent and substantial improvements for Soft-CFT over Hard-CFT. Predictors trained on counterfactuals generated with our Soft-CFT strategy are much closer to baseline performance on both intervened and unintervened attributes for all three sets of race, sex, and disease counterfactuals. Hard-CFT, on the other hand, appears to introduce undesired distribution shift. Note the decrease in performance on sex counterfactuals even for the sex predictor, despite their high effectiveness observed in \cref{tab:effectiveness}. Here, attribute amplification of Hard-CFT results in images being too extreme. While they are correctly classified with predictors trained on real images, these images are not suitable for training predictors evaluated on real images. The results in \cref{tab:downstream_predictors} together with \cref{tab:effectiveness} indicate that Soft-CFT both improves effectiveness and reduces attribute amplification yielding more faithful counterfactual images which is important for their utility in downstream applications.


\begin{table}[t!]
\centering
\caption{Performance of predictors trained on CFs  with \interv{intervened} and \unint{unintervened} attributes, evaluated on real images. Comparing different CF training strategies. AUC $\%$ performance is compared to a baseline trained on real images.
}
\label{tab:downstream_predictors}
\begin{tabular}{@{}lp{0.03\textwidth}lp{0.02\textwidth}cp{0.03\textwidth}cp{0.03\textwidth}c@{}}
\toprule
\multirow{2}{*}{\textsc{Training Set}} & & \multirow{2}{*}{\textsc{CF Type}} && \multicolumn{5}{c}{\textsc{Attribute Prediction Task}} \\
\cmidrule{5-9}
  & & && \textsc{Race}        && \textsc{Sex}         && \textsc{Disease}       \\ 
\midrule
Original   &      &  Real images    && 93.8        && 99.6        && 94.2          \\

\midrule
\multirow{3}{*}{Race CFs}  && No CFT& & \cellcolor{graybg} \interv{56.9 \small{($\downarrow$ 36.9)}}    && \unint{99.5 \small{($\downarrow$ 0.1)}}     && \unint{93.9 \small{($\downarrow$ 0.3)}}    \\
&& Hard-CFT && \cellcolor{graybg} \interv{81.6 \small{($\downarrow$  12.2)}}       && \unint{98.3 \small{($\downarrow$ 1.3)}}     && \unint{92.2 \small{($\downarrow$ 2.0)}}       \\
&& Soft-CFT  && \cellcolor{graybg} \interv{81.2 \small{($\downarrow$ 12.6)}}     && \unint{99.6 \small{($\cdot$ 0.0)}}     && \unint{93.8 \small{($\downarrow$  0.4)}}        \\
\midrule
\multirow{3}{*}{Sex CFs} && No CFT && \unint{82.1 \small{($\downarrow$ 11.7)}}     && \cellcolor{graybg} \interv{82.2 \small{($\downarrow$ 17.4)}}      && \unint{92.3 \small{($\downarrow$ 1.9)}}         \\
&& Hard-CFT && \unint{75.4 \small{($\downarrow$ 18.4)}}      && \cellcolor{graybg} \interv{79.6 \small{($\downarrow$ 20.0)}}      && \unint{89.8 \small{($\downarrow$  4.4)}}          \\
&& Soft-CFT && \unint{92.4 \small{($\downarrow$ 1.4)}}      && \cellcolor{graybg} \interv{90.0 \small{($\downarrow$ 9.6)}}   && \unint{94.0 \small{($\downarrow$ 0.2)}}         \\
\midrule
\multirow{3}{*}{Disease CFs} && No CFT     && \unint{79.7 \small{($\downarrow$  14.1)}}     && \unint{98.1 \small{($\downarrow$ 1.5)}}     && \cellcolor{graybg} \interv{30.4 \small{($\downarrow$  63.8)}}      \\
&& Hard-CFT && \unint{77.6 \small{($\downarrow$  16.2)}}       && \unint{95.2 \small{($\downarrow$  4.4)}}    && \cellcolor{graybg} \interv{88.8 \small{($\downarrow$  5.4)}}      \\
&& Soft-CFT && \unint{91.8 \small{($\downarrow$  2.0)}}       && \unint{99.5 \small{($\downarrow$  0.1)}}       && \cellcolor{graybg} \interv{90.0 \small{($\downarrow$  4.2)}}          \\ 
\bottomrule   
\end{tabular}
\end{table}

\subsection{Assessing distribution shift of counterfactual images}
To gain further insights into the faithfulness of counterfactual images under different training strategies, we inspect the latent space of image embeddings extracted from attribute predictors trained on real data. Here, we use a publicly available pre-trained multi-task model from~\citet{glocker2023algorithmic}. This model was trained on the same splits of MIMIC-CXR as used in our work. The model was previously used to visualise subgroup differences across race, sex and disease. Applying principal component analysis (PCA) to the high-dimensional image embeddings showed that each independent attribute is encoded in a different PCA mode, with the first mode encoding disease, the second mode encoding sex, and the third mode encoding race differences. We leverage this feature inspection approach to compare image embeddings from real and counterfactual images. Ideally, we should be unable to see differences in the data distribution between real and corresponding generated images, if the counterfactuals are faithful.

\begin{figure*}[!t]
     \centering
          \begin{subfigure}{0.49\textwidth}
         \centering
         \includegraphics[width=\textwidth]{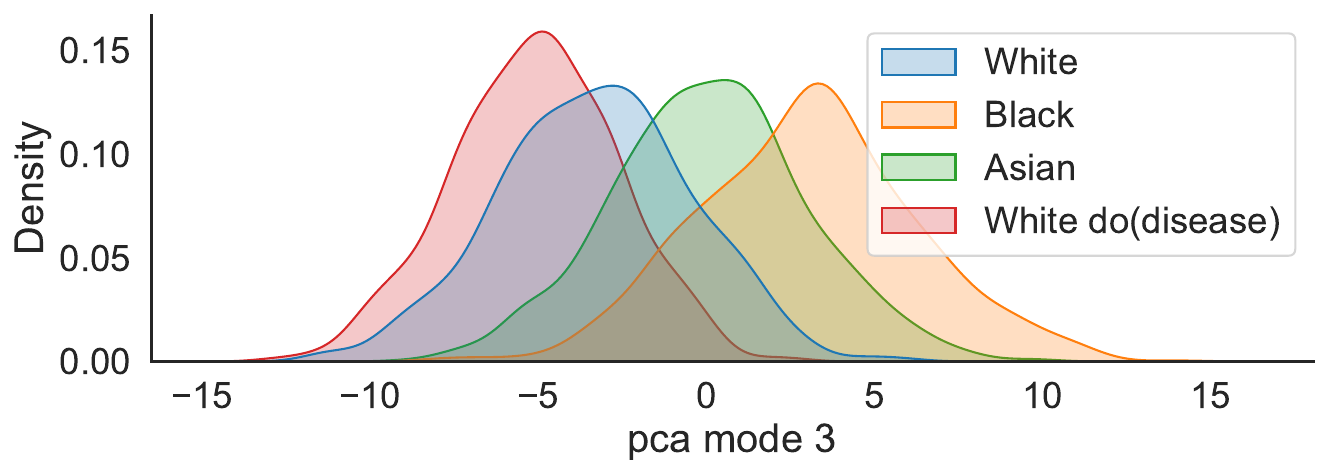}
     \end{subfigure}
     \hfill
     \begin{subfigure}{0.49\textwidth}
         \centering
         \includegraphics[width=\textwidth]{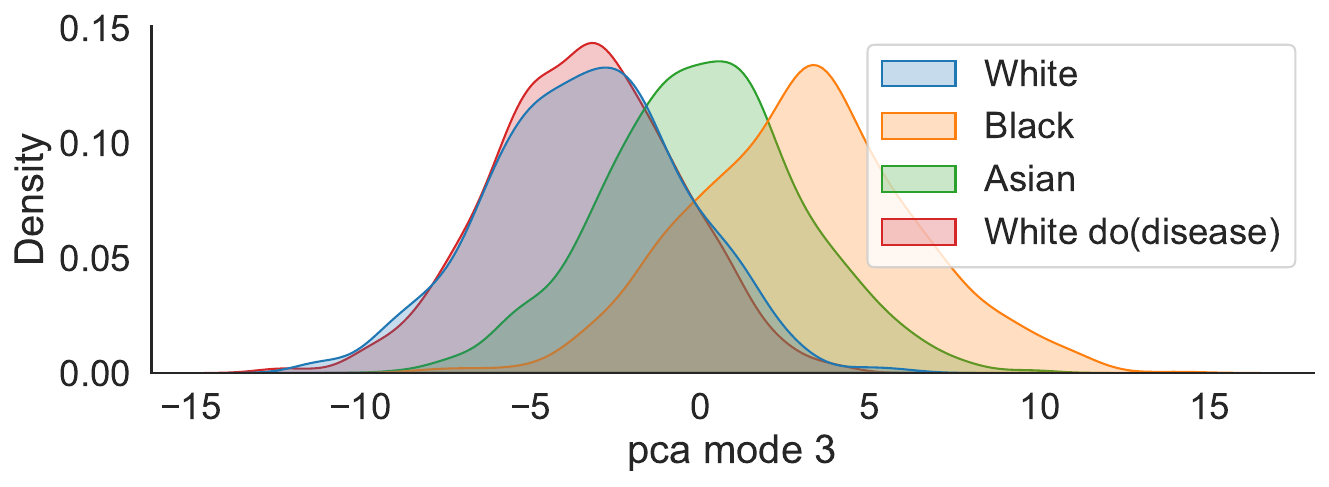}
     \end{subfigure}
     
     \begin{subfigure}{0.49\textwidth}
         \centering
         \includegraphics[width=\textwidth]{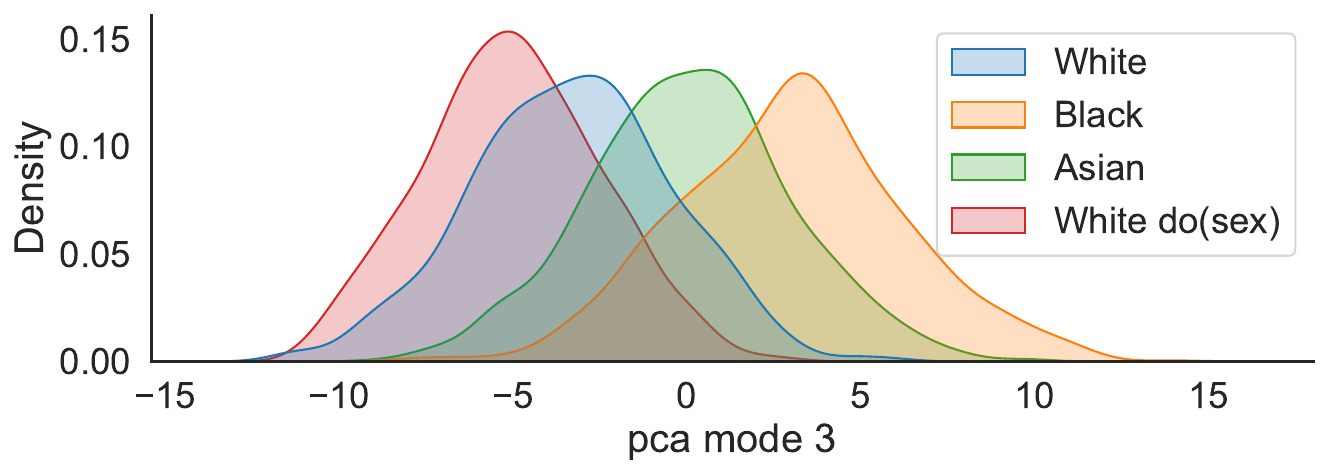}
         \caption{Hard-CFT}
     \end{subfigure}
     \hfill
     \begin{subfigure}{0.49\textwidth}
         \centering
         \includegraphics[width=\textwidth]{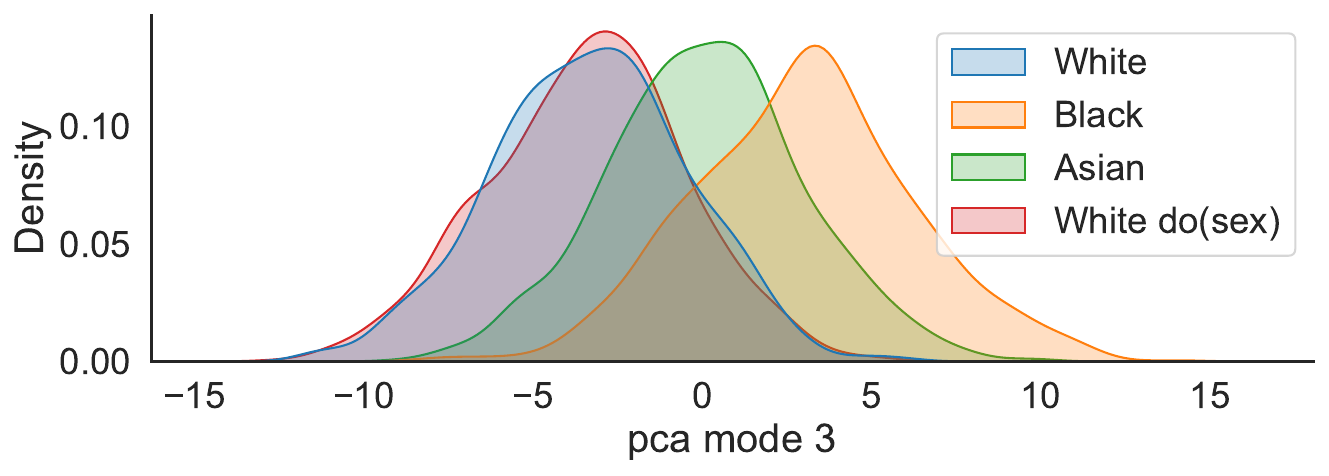}
          \caption{Soft-CFT}
     \end{subfigure}
    %
     
     %
\caption{Marginal distribution of image embeddings across PCA mode 3 of a multi-task model. This mode encodes changes in the race attribute. We plot distributions for subgroups of real data alongside distributions of counterfactual images when intervening on disease and sex attributes. When training with Hard-CFT (left) there is a clear distribution shift between real (blue) and counterfactual images (red). This shift is removed for both interventions when using our proposed Soft-CFT (right). These results suggest that Soft-CFT successfully mitigates attribute amplification and generates more faithful counterfactual images.}
\label{fig: pca}
\end{figure*}

We randomly select 1000 samples from the test set, generate counterfactuals and inspect the distributions in the corresponding PCA modes. \cref{fig: pca} focuses on distribution shift in the race attribute under interventions on disease and sex. We observe that with Hard-CFT, there is a clear distribution shift between the real images from the subgroup `White' and the counterfactual subgroups `White $do(\textrm{disease})$' and `White $do(\textrm{sex})$'. In contrast, the images generated with Soft-CFT align much better with real data distribution, suggesting that Soft-CFT mitigates attribute amplification and yields more faithful counterfactual images that are more similar to real data. Additional plots inspecting other combinations of intervened and unintervened attributes across different subgroups are provided in \cref{fig:additional_pca}, confirming that Soft-CFT improves counterfactual image generation.

\section{Conclusion}
In this paper, we discover the important issue of \textit{attribute amplification} in counterfactual image generation with DSCMs when using the previously proposed counterfactual fine-tuning step with hard labels~\citep{de2023high}. Attribute amplification violates the assumed causal graph, introducing distribution shift, and resulting in potentially harmful spurious correlations. For example, we observed how interventions on independent attributes such as race and sex can cause distribution shift in disease status. An intervened image may appear healthier or more diseased than the original real images. Such undesired effects could be harmful when counterfactual images are used for data augmentation in downstream applications. We mitigate this with Soft-CFT, using inferred probabilities for unintervened attributes. Our experiments on a large chest X-ray dataset with results on attribute prediction performance, together with a detailed feature inspection to analyse distribution shift across subgroups, suggest that Soft-CFT successfully mitigates attribute amplification with the ability to generate effective and more faithful counterfactual images. However, we still observe discrepancies between the attribute prediction performance when training on real vs counterfactual images. Closing this gap remains an open problem for future work.

The proposed method is not only applicable to DSCMs but may improve other causal generative models that use classifier guidance during training. Thus, our work makes an important advancement towards more faithful and hopefully unbiased causal modelling in medical imaging. All code will be publicly released.

\section*{Acknowledgements}
We acknowledge support of the UKRI AI programme, and the Engineering and Physical Sciences Research Council (EPSRC), for CHAI - EPSRC AI Hub for Causality in Healthcare AI with Real Data (grant number EP/Y028856/1). T.X., F.R. and B.G. also received funding from the European Research Council (ERC) under the European Union’s Horizon 2020 research and innovation programme (grant agreement No 757173, project MIRA, ERC-2017-STG). B.G. also received support from the Royal Academy of Engineering as part of his Kheiron/RAEng Research Chair in Safe Deployment of Medical Imaging AI. M.R. is funded through an Imperial College London President's PhD Scholarship. C.J. is supported by Microsoft Research, EPSRC and The Alan Turing Institute through a Microsoft PhD Scholarship and Turing PhD enrichment award.

\bibliographystyle{tmlr}
\bibliography{ref}

\newpage
\appendix
\setcounter{table}{0}
\renewcommand{\thetable}{\Alph{section}\arabic{table}}
\setcounter{figure}{0}
\renewcommand{\thefigure}{\Alph{section}\arabic{figure}}

\section{Supplementary Material}

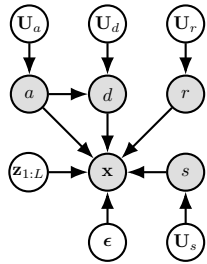
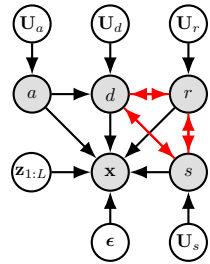
\begin{figure*}[h!]
\centering
    \hfill
        \begin{subfigure}{.45\textwidth}
        \centering
        \begin{tikzpicture}[thick]
            \node[obs,     scale=0.70] (x) {$\mathbf{x}$};
            \node[obs,      scale=0.70, above=15pt of x] (d) {$d$};
            \node[obs,      scale=0.70, right=15pt of d] (r) {$r$};
            \node[obs,      scale=0.70, left=15pt of d] (a) {$a$};
            \node[obs,      scale=0.70, right=15pt of x] (s) {$s$};

            \node[latent,      scale=0.70, xshift=0pt, left=15pt of x] (zx) {$\mathbf{z}_{1:L}$};
            \node[latent,      scale=0.70, xshift=0pt, below=12pt of x] (ux) {$\boldsymbol{\epsilon}$};
            \node[latent,      scale=0.70, below=12pt of s] (us) {$\mathbf{U}_s$};
            \node[latent,      scale=0.70, above=12pt of r] (ur) {$\mathbf{U}_r$};
            \node[latent,      scale=0.70, above=12pt of d] (ud) {$\mathbf{U}_d$};
            \node[latent,      scale=0.70, above=12pt of a] (ua) {$\mathbf{U}_a$};

            \edge[-{Latex[scale=1.0]}]{a}{x}
            \edge[-{Latex[scale=1.0]}]{a}{d}
            \edge[-{Latex[scale=1.0]}]{d}{x}
            \edge[-{Latex[scale=1.0]}]{r}{x}
            \edge[-{Latex[scale=1.0]}]{s}{x}

            \draw [-{Latex[scale=1.0]}] (us) to [out=90,in=-90] (s);
            \draw [-{Latex[scale=1.0]}] (ua) to [out=-90,in=90] (a);
            \draw [-{Latex[scale=1.0]}] (ur) to [out=-90,in=90] (r);
            \draw [-{Latex[scale=1.0]}] (ud) to [out=-90,in=90] (d);
            \draw [-{Latex[scale=1.0]}] (ux) to (x);
            \draw [-{Latex[scale=1.0]}] (zx) to (x);
        \end{tikzpicture}
        \hfill
        \caption{Assumed causal graph for MIMIC-CXR.  Variables in the causal graph are: age $(a)$, sex $(s)$, race $(r)$, disease $(d)$ (pleural effusion) and chest X-ray $(\mathbf{x})$. }
        \label{fig:scm_chest}
    \end{subfigure}
\hfill
    \begin{subfigure}{.45\textwidth}
    \centering
        \begin{tikzpicture}[thick]
            \node[obs,     scale=0.70] (x) {$\mathbf{x}$};
            \node[obs,     scale=0.70, above=15pt of x] (d) {$d$};
            \node[obs,     scale=0.70, right=15pt of d] (r) {$r$};
            \node[obs,     scale=0.70, left=15pt of d] (a) {$a$};
            \node[obs,     scale=0.70, right=15pt of x] (s) {$s$};

            \node[latent,     scale=0.70, xshift=0pt, left=15pt of x] (zx) {$\mathbf{z}_{1:L}$};
            \node[latent,  scale=0.70, xshift=0pt, below=12pt of x] (ux) {$\boldsymbol{\epsilon}$};
            \node[latent,     scale=0.70, below=12pt of s] (us) {$\mathbf{U}_s$};
            \node[latent,     scale=0.70, above=12pt of r] (ur) {$\mathbf{U}_r$};
            \node[latent,     scale=0.70, above=12pt of d] (ud) {$\mathbf{U}_d$};
            \node[latent,     scale=0.70, above=12pt of a] (ua) {$\mathbf{U}_a$};

            \edge[-{Latex[scale=1.0]}]{a}{x}
            \edge[-{Latex[scale=1.0]}]{a}{d}
            \edge[-{Latex[scale=1.0]}]{d}{x}
            \edge[-{Latex[scale=1.0]}]{r}{x}
            \edge[-{Latex[scale=1.0]}]{s}{x}
            \edge[-{Latex[scale=1.0]}, red]{s}{d}
            \edge[-{Latex[scale=1.0]}, red]{d}{s}
            \edge[-{Latex[scale=1.0]}, red]{s}{r}
            \edge[-{Latex[scale=1.0]}, red]{r}{s}
            \edge[-{Latex[scale=1.0]}, red]{r}{d}
            \edge[-{Latex[scale=1.0]}, red]{d}{r}
            

            \draw [-{Latex[scale=1.0]}] (us) to [out=90,in=-90] (s);
            \draw [-{Latex[scale=1.0]}] (ua) to [out=-90,in=90] (a);
            \draw [-{Latex[scale=1.0]}] (ur) to [out=-90,in=90] (r);
            \draw [-{Latex[scale=1.0]}] (ud) to [out=-90,in=90] (d);
            \draw [-{Latex[scale=1.0]}] (ux) to (x);
            \draw [-{Latex[scale=1.0]}] (zx) to (x);
        \end{tikzpicture}
        \caption{With attribute amplification, the assumed causal graph is violated. For instance, $do(s)$ affecting \textit{d} could make \textit{s} a parent of \textit{d} and vice versa.  }
        \label{fig:violated_scm}
    \end{subfigure}
    \caption{Illustration of how attribute amplification may violate the causal graph pre-defined for the DSCM which may lead to spurious correlations between protected characteristics and disease status encoded in the counterfactual images. }
    \label{fig:amplification violate causal graph}
\end{figure*}

\begin{figure*}[h!]
    \centering
    \begin{subfigure}{0.566\textwidth}
    \centering
    \includegraphics[width=\textwidth]{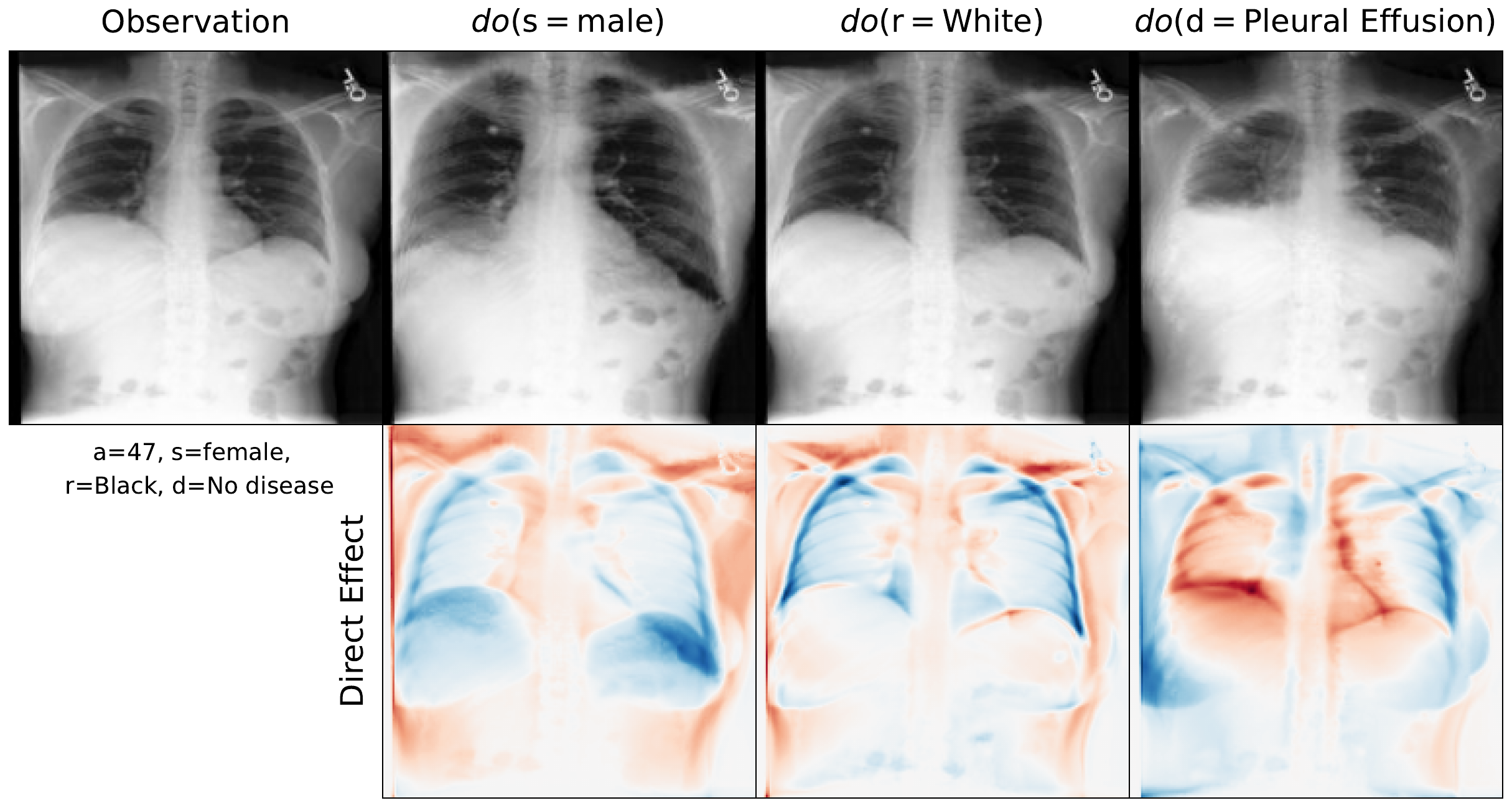}
    \end{subfigure}
    \begin{subfigure}{0.424\textwidth}
    \centering
    \includegraphics[width=\textwidth,trim={10.35cm 0 0 0},clip]{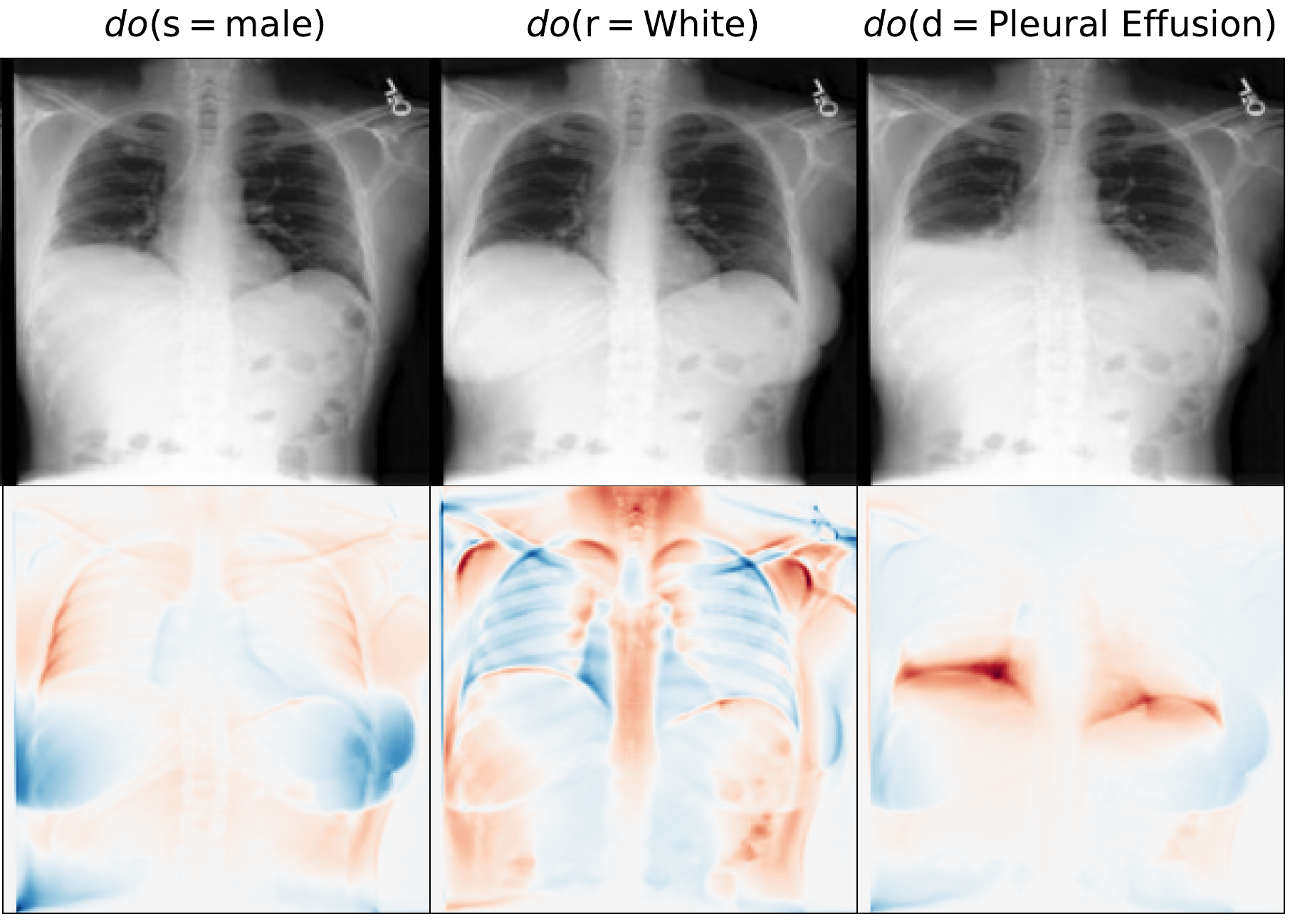}
    \end{subfigure}
    
    \centering
    \begin{subfigure}{0.566\textwidth}
    \centering
    \includegraphics[width=\textwidth]{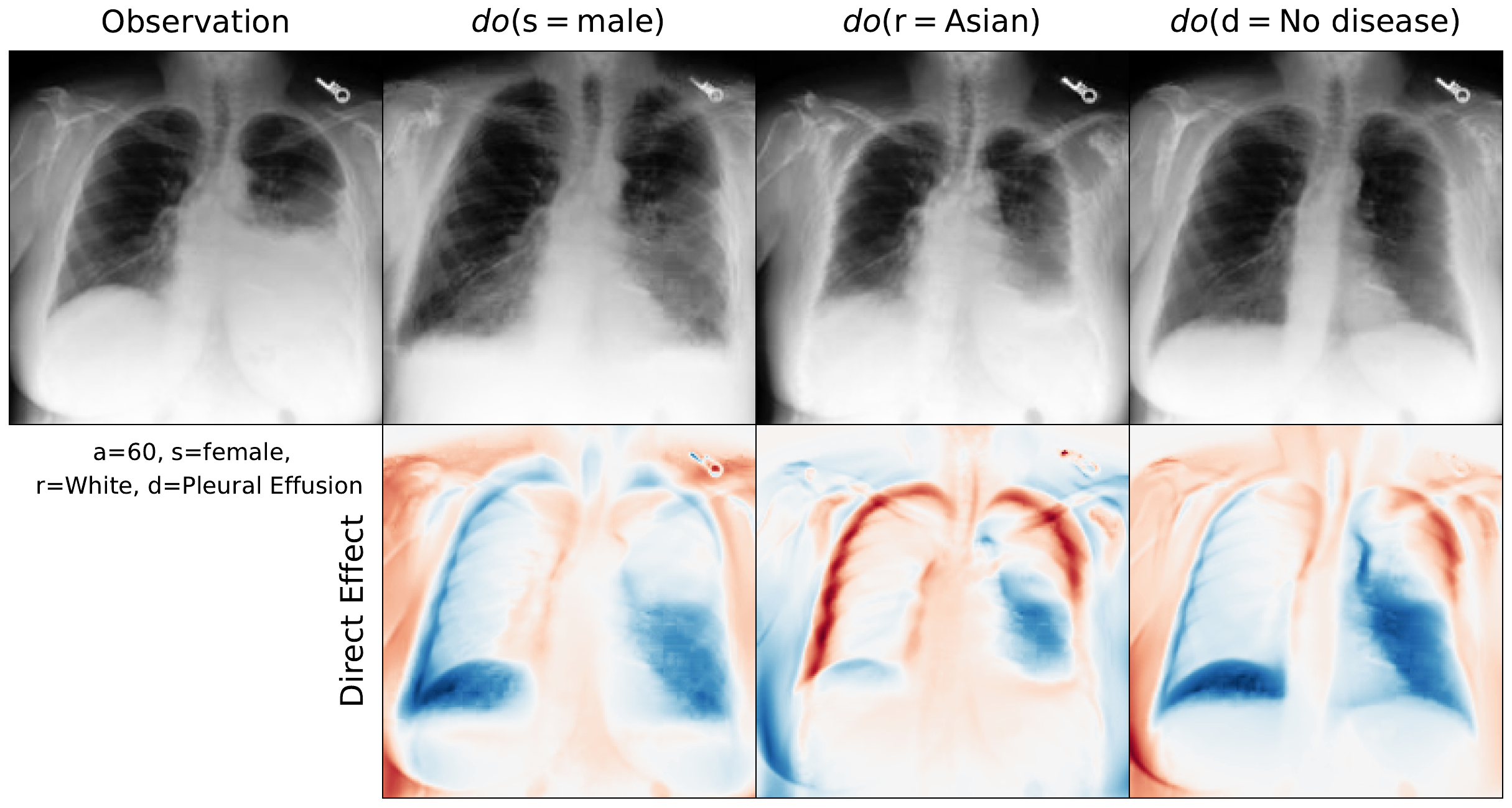}
    \caption{Hard-CFT}
    \end{subfigure}
    \begin{subfigure}{0.424\textwidth}
    \centering
    \includegraphics[width=\textwidth,trim={10.35cm 0 0 0},clip]{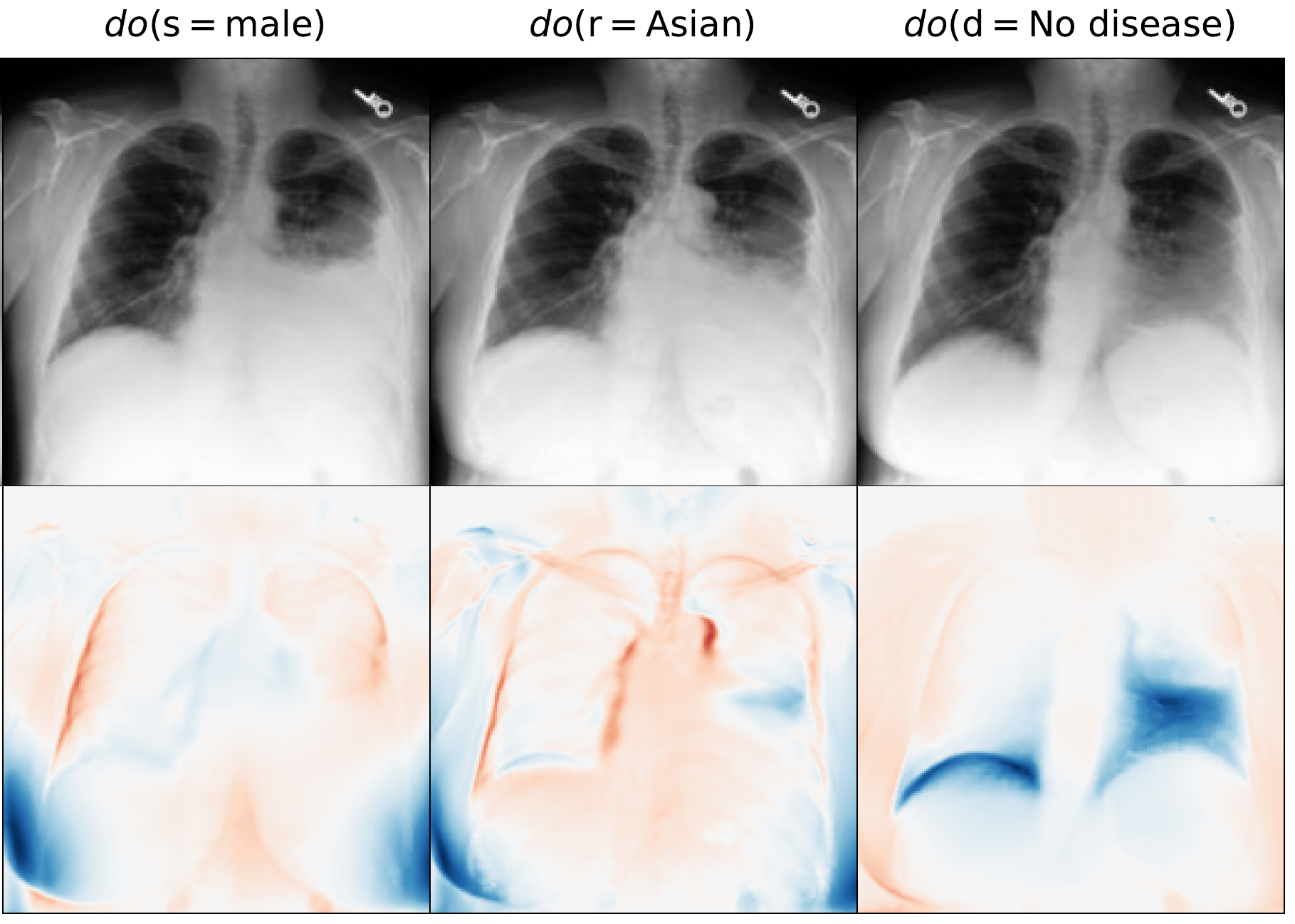}
    \caption{Soft-CFT}
    \end{subfigure}
    \caption{Generated CFs with (a) Hard-CFT and (b) Soft-CFT. Top rows show original image $\mathbf{x}$ and CFs $\mathbf{\widetilde{x}}$; bottom rows show direct effect of CFs, i.e. $\mathbf{\widetilde{x}}-\mathbf{x}$. }
    \label{fig:more CFs}
\end{figure*}

\begin{figure*}[t!]
 \begin{subfigure}{0.40\textwidth}
         \centering         \includegraphics[width=\textwidth]{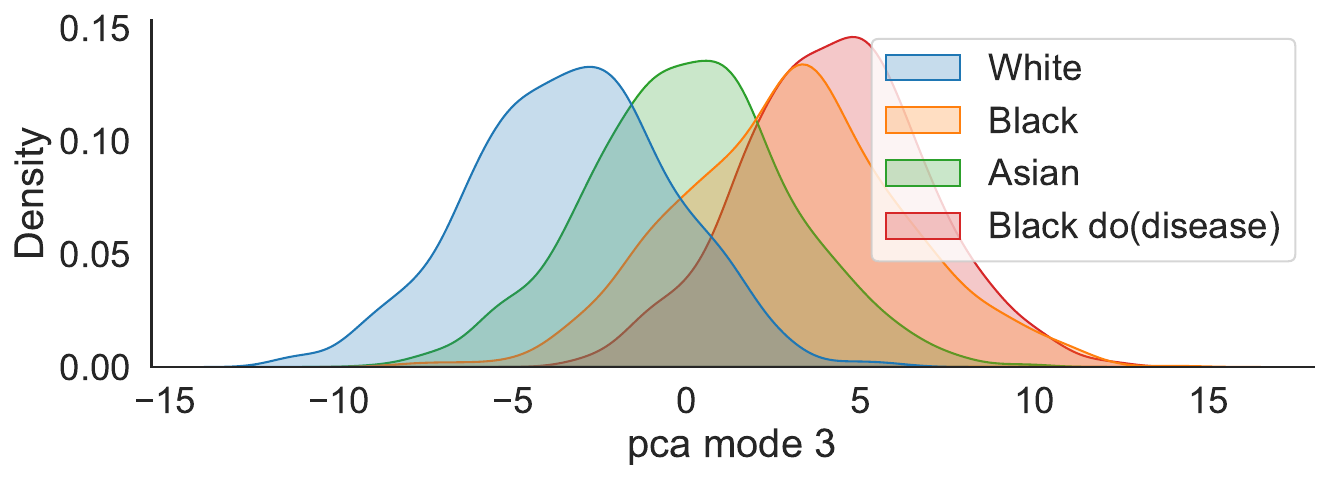}
     \end{subfigure}
     \hfill
     \begin{subfigure}{0.40\textwidth}
         \centering
         \includegraphics[width=\textwidth]{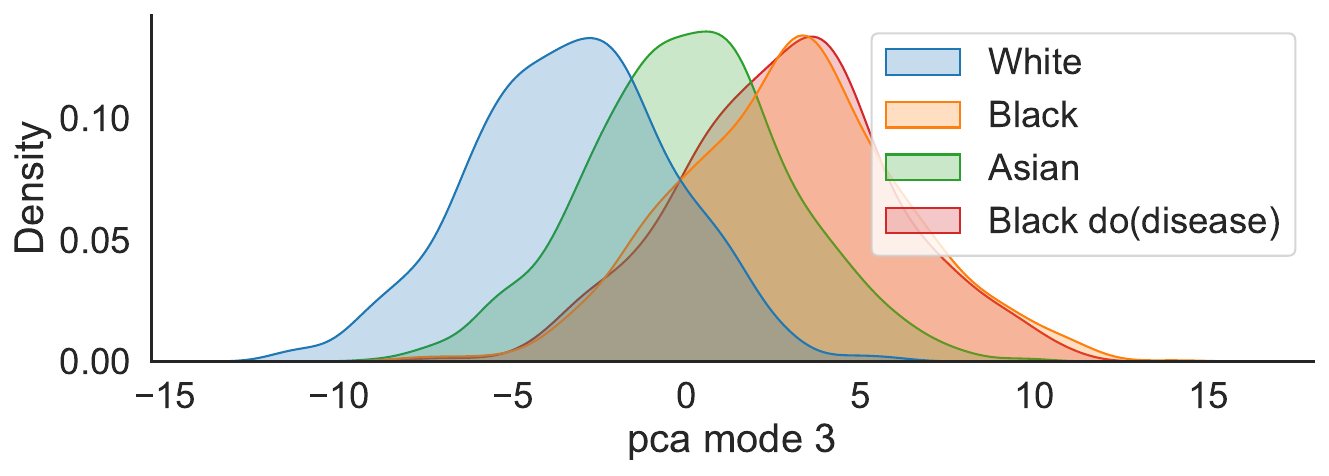}
     \end{subfigure}
 \begin{subfigure}{0.40\textwidth}
         \centering         \includegraphics[width=\textwidth]{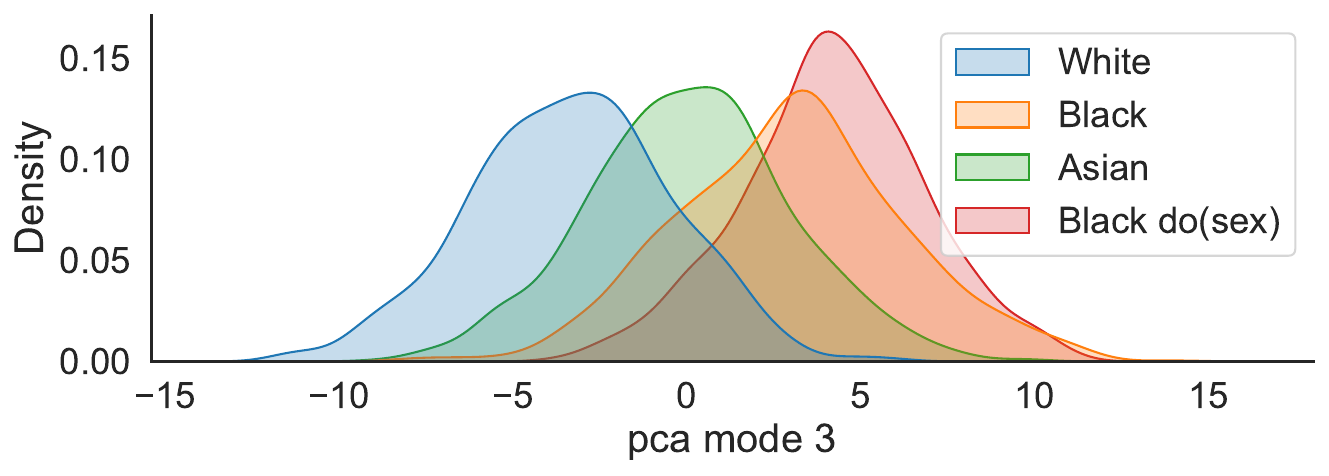}
     \end{subfigure}
     \hfill
     \begin{subfigure}{0.40\textwidth}
         \centering
         \includegraphics[width=\textwidth]{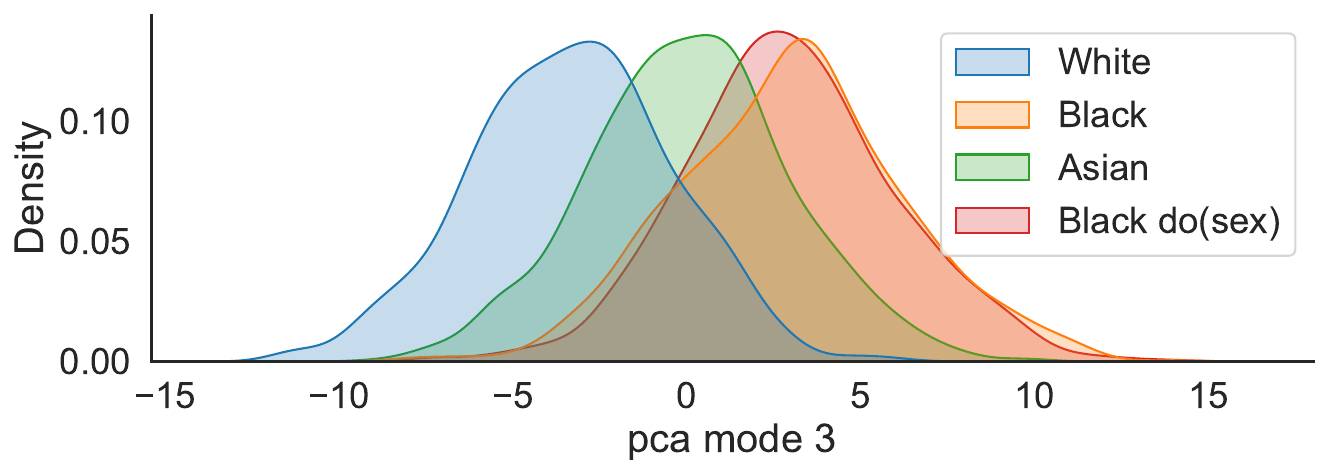}
     \end{subfigure}
 \begin{subfigure}{0.40\textwidth}
         \centering         \includegraphics[width=\textwidth]{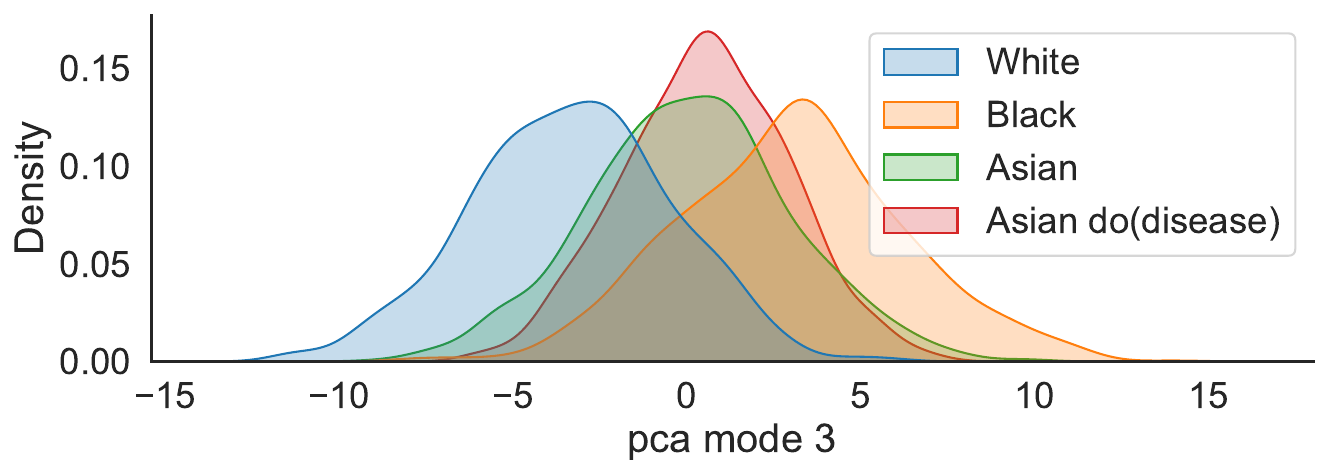}
     \end{subfigure}
     \hfill
     \begin{subfigure}{0.40\textwidth}
         \centering
         \includegraphics[width=\textwidth]{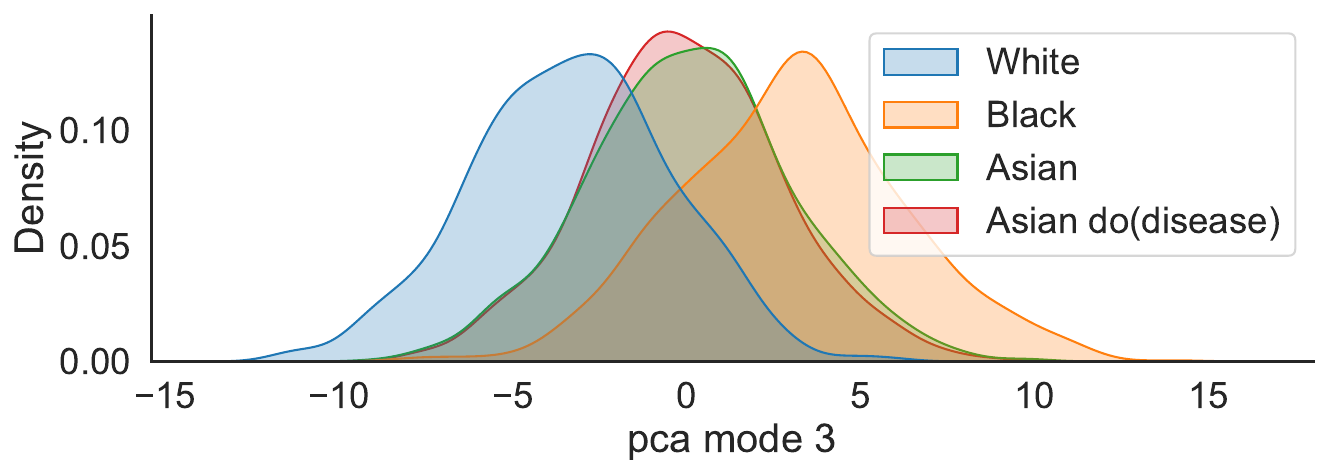}
     \end{subfigure}
 \begin{subfigure}{0.40\textwidth}
         \centering         \includegraphics[width=\textwidth]{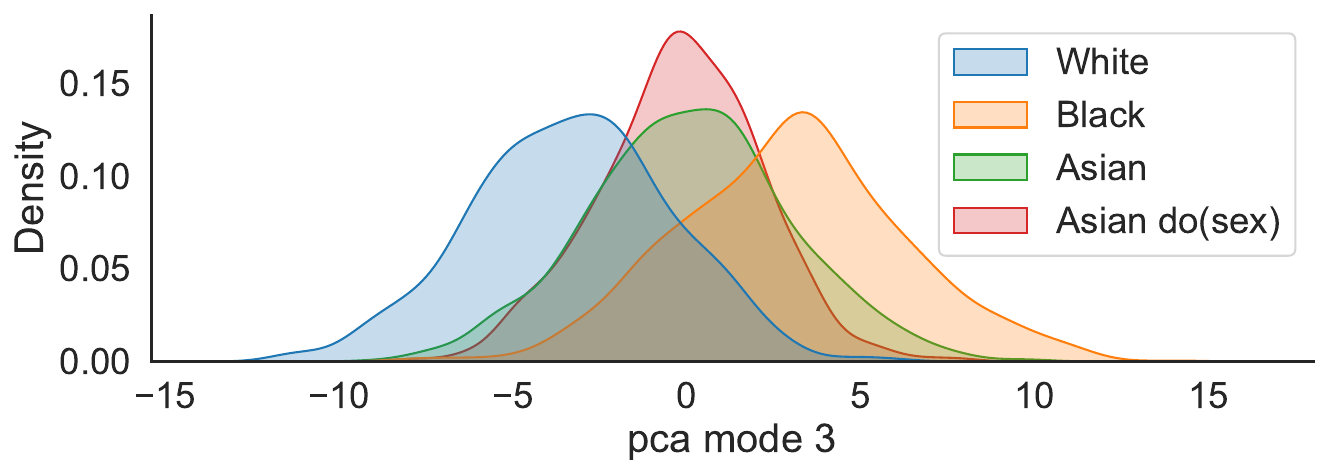}
     \end{subfigure}
     \hfill
     \begin{subfigure}{0.40\textwidth}
         \centering
         \includegraphics[width=\textwidth]{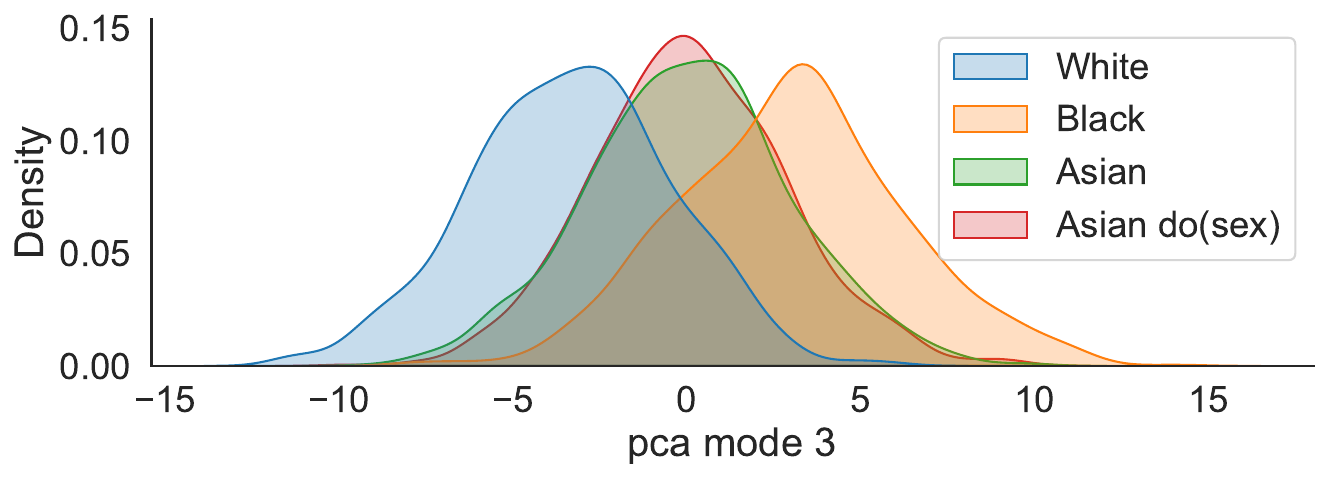}
     \end{subfigure}
 \begin{subfigure}{0.40\textwidth}
         \centering         \includegraphics[width=\textwidth]{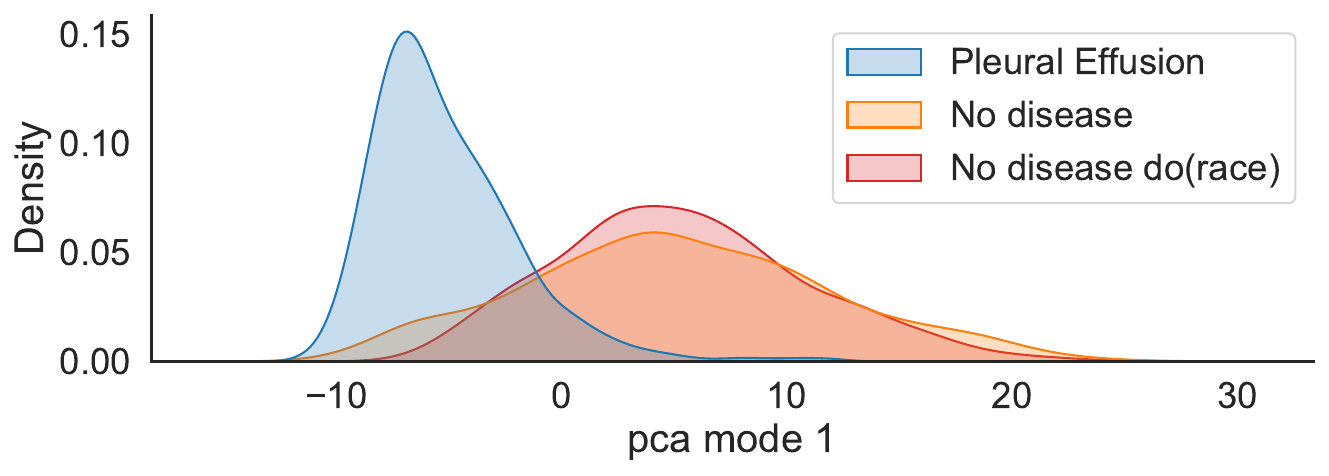}
     \end{subfigure}
     \hfill
     \begin{subfigure}{0.40\textwidth}
         \centering
         \includegraphics[width=\textwidth]{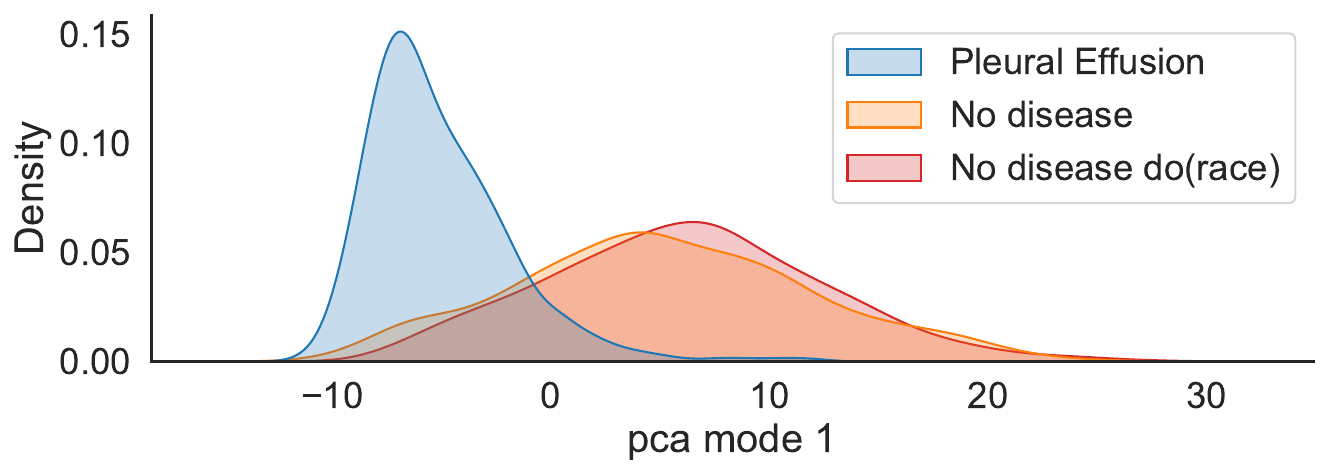}
     \end{subfigure}
 \begin{subfigure}{0.40\textwidth}
         \centering         \includegraphics[width=\textwidth]{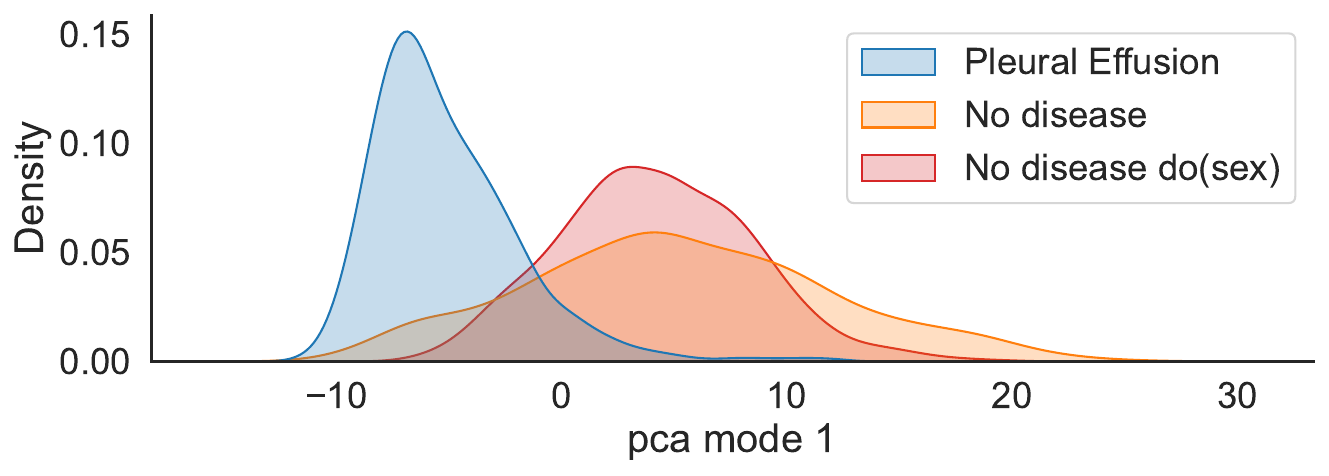}
     \end{subfigure}
     \hfill
     \begin{subfigure}{0.40\textwidth}
         \centering
         \includegraphics[width=\textwidth]{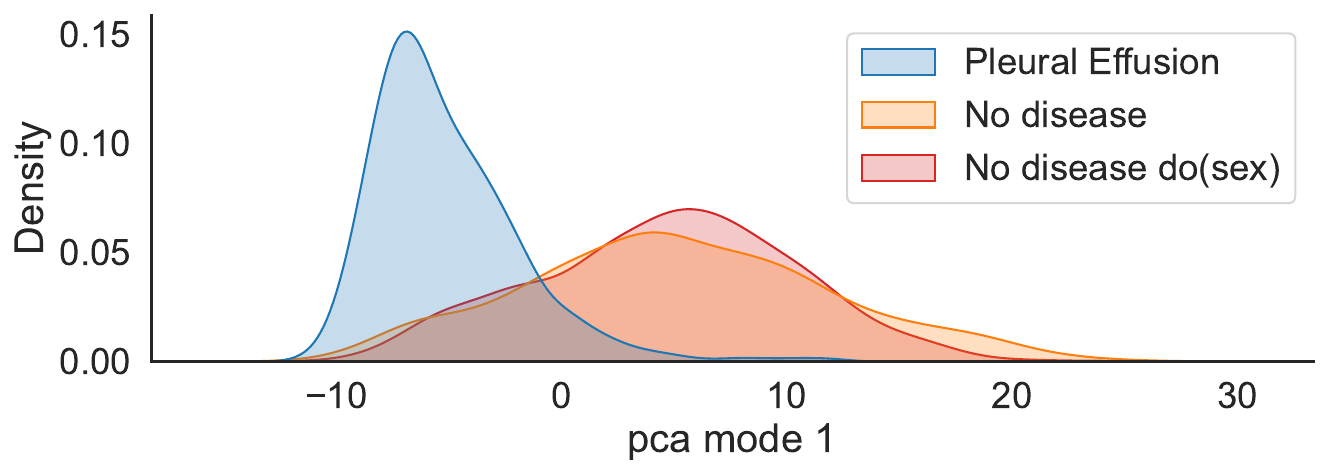}
     \end{subfigure}
 \begin{subfigure}{0.40\textwidth}
         \centering         \includegraphics[width=\textwidth]{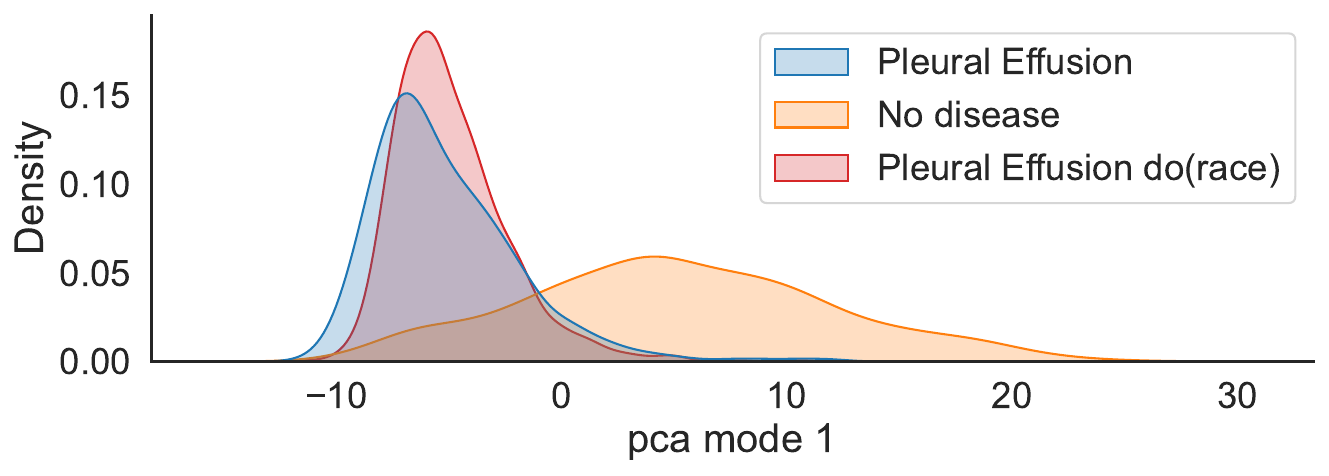}
     \end{subfigure}
     \hfill
     \begin{subfigure}{0.40\textwidth}
         \centering
         \includegraphics[width=\textwidth]{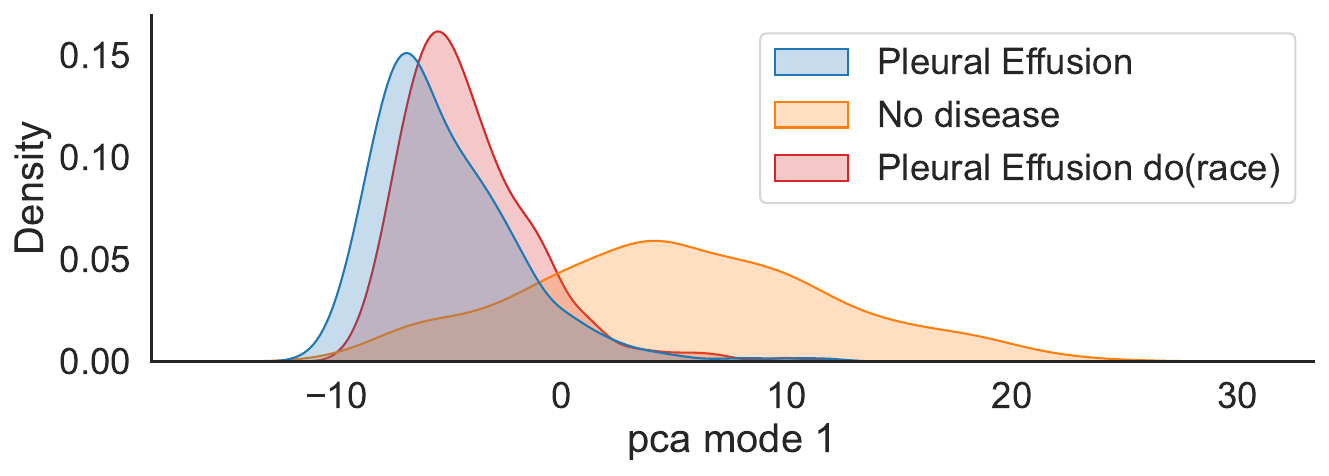}
     \end{subfigure}
 \begin{subfigure}{0.40\textwidth}
         \centering         \includegraphics[width=\textwidth]{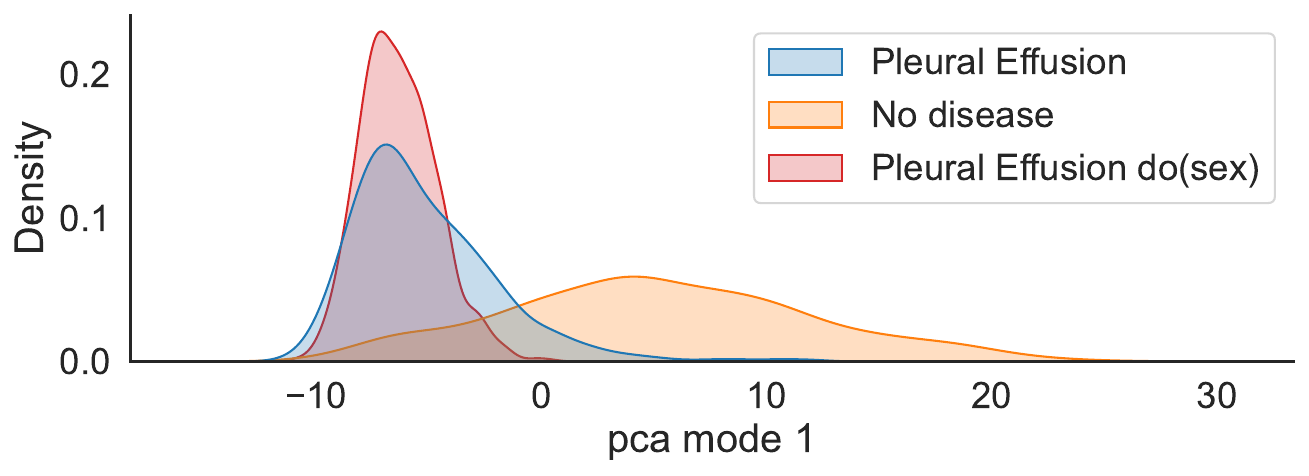}
        \caption{Hard-CFT}
     \end{subfigure}
     \hfill
     \begin{subfigure}{0.40\textwidth}
         \centering
         \includegraphics[width=\textwidth]{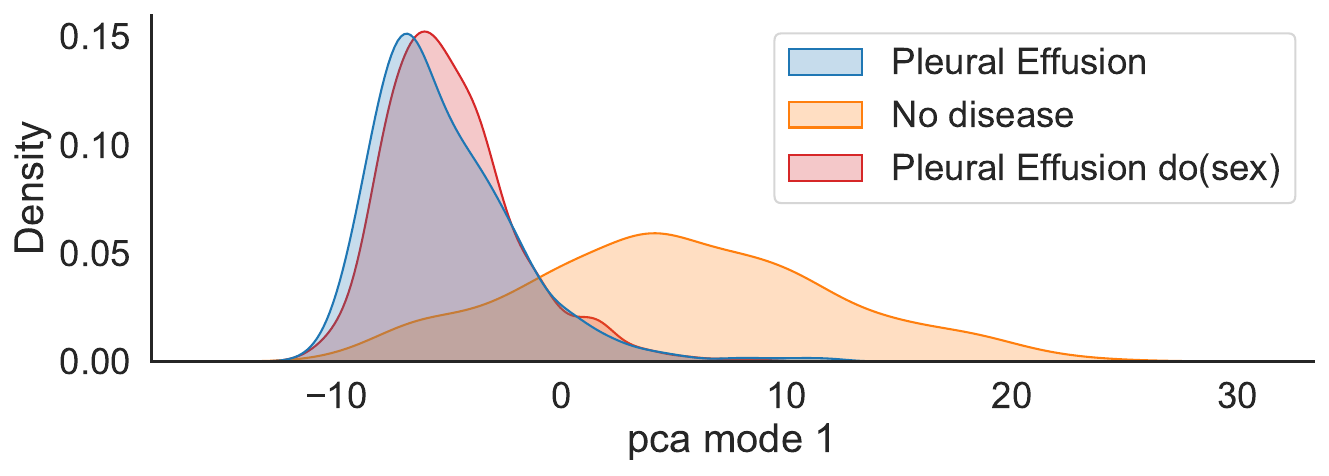}
         \caption{Soft-CFT}
     \end{subfigure}
\caption{Marginal distribution of PCA modes of pretrained embeddings from a multi-task model predicted all attributes. We plot embeddings of real data along side with generated counterfactuals of various subgroups. We can see that when training with Hard-CFT (left) there is a distribution shift between real images and images after intervention (red). Conversely, this shift is mitigated when using our proposed soft counterfactual fine-tuning (Soft-CFT, right).}
\label{fig:additional_pca}
\end{figure*}
\end{document}